\documentclass[runningheads]{llncs}
\usepackage[T1]{fontenc}
\usepackage[utf8]{inputenc}
\usepackage{graphicx,verbatim}
\usepackage{silence}
\usepackage{amsmath,amssymb}
\usepackage{multirow}
\usepackage{subcaption}
\usepackage{placeins}

\usepackage{bm} % For bold math symbols
\usepackage{cite} % Referansları [1,2] gibi düzenli gösterir
\usepackage{float}
\begin{document}
\title{SCDM: Spatial-Contextual Disentanglement Mamba via Differential Inference for Efficient Image Classification}
\titlerunning{SCDM: Disentanglement Mamba for Image Classification}
\begin{comment}  %% Removed for anonymized MICCAI submission
\author{First Author\inst{1}\orcidID{0000-1111-2222-3333} \and
Second Author\inst{2,3}\orcidID{1111-2222-3333-4444} \and
Third Author\inst{3}\orcidID{2222--3333-4444-5555}}
%
\authorrunning{F. Author et al.}
\institute{Princeton University, Princeton NJ 08544, USA \and
Springer Heidelberg, Tiergartenstr. 17, 69121 Heidelberg, Germany
\email{lncs@springer.com}\\
\url{http://www.springer.com/gp/computer-science/lncs} \and
ABC Institute, Rupert-Karls-University Heidelberg, Heidelberg, Germany\\
\email{\{abc,lncs\}@uni-heidelberg.de}}

\end{comment}

\author{Mustafa Bora Çelik\inst{1} \and Hayriye Aktaş Dinçer\inst{1} \and Ayse Keles\inst{2}}
\authorrunning{M. B. Çelik et al.}
\institute{
  Department of Electrical and Electronics Engineering, Ankara Medipol University, Ankara, Turkiye \\
  \email{mustafa.celik1@std.ankaramedipol.edu.tr, hayriye.aktas@ankaramedipol.edu.tr} \and
  School of Computer Science, University of Galway, Galway, Ireland \\
  \email{ayse.keles@universityofgalway.ie}
}
  
\maketitle              % typeset the header of the contribution
\begin{abstract}
State Space Models (SSMs), particularly VMamba, have emerged as efficient alternatives for modeling long-range dependencies in medical image analysis. However, distinguishing subtle pathological features from visually similar anatomical backgrounds remains a significant challenge. Existing SSM architectures often learn entangled representations, lacking explicit mechanisms to separate disease-specific signals from normal anatomy. To address this limitation, we propose Spatial-Contextual Differential Mamba (SCDM), an asymmetric dual-branch architecture designed for selective representational disentanglement. SCDM introduces a Positive Branch for extracting discriminative features and a Negative Branch that actively models and suppresses normal anatomical context. This separation is achieved through a similarity-driven repulsion gate and a differential inference rule, which promote competitive feature learning without requiring additional branch labels or increasing model capacity. Evaluated on the RSNA Pneumonia dataset, SCDM achieves competitive classification performance (AUC of 0.858) while requiring significantly fewer parameters (29.4M) and FLOPs (1.44G) compared to standard VMamba and vision transformer baselines. Furthermore, activation analyses demonstrate that our differential mechanism yields highly precise localization, effectively isolating lesions by inhibiting irrelevant anatomical distractors.

% Authors must provide keywords and are not allowed to remove this Keyword section.

\end{abstract}

\section{Introduction}
Medical image classification is a cornerstone of computer-aided diagnosis, enabling early detection of various pathologies \cite{liu2023lie, neelima2024survey, zhou2021review}. Over the last decade, Convolutional Neural Networks (CNNs) such as ResNet \cite{he2016resnet} have set competitive benchmarks by leveraging local inductive biases. However, their fixed kernel size inherently limits the ability to capture long-range global context, which is often crucial for identifying subtle pathological patterns across high-resolution scans \cite{raghu2019transfusion}. Vision Transformers (ViTs) \cite{dosovitskiy2021image} and their hierarchical variants like Swin Transformers \cite{liu2021swin} overcame this by utilizing self-attention for global spatial modeling. Yet, the quadratic computational complexity of attention mechanisms remains a significant bottleneck for real-time clinical deployment. To address this, State Space Models (SSMs) like Mamba have emerged as efficient alternatives, offering linear-complexity modeling of long-range context \cite{gu2023mamba, liu2024vmamba}.

Despite these advances, a persistent challenge in medical imaging is the entanglement of pathological features with visually similar anatomical backgrounds \cite{chen2022recent}. Standard architectures often lack specialized mechanisms to filter out irrelevant anatomical context (e.g., bony structures in chest X-rays), which can lead to false positives. To overcome this, we propose Spatial-Contextual Differential Mamba (SCDM), an asymmetric dual-branch architecture tailored for representational disentanglement. SCDM integrates a Positive Branch to extract discriminative features and an active Negative Branch capable of modeling and suppressing normal anatomical mimics. Through a similarity-driven repulsion mechanism and differential inference, SCDM isolates pathology-relevant signals without requiring increased model complexity and parameters.

\section{Preliminaries}
In this work, we adopt the standard VMamba backbone and introduce a differential dual-branch extension at the SS2D level, enabling explicit supervision and disentanglement of hidden state representations without altering the overall backbone structure.

\subsubsection{Selective State Space Modeling.} State Space Models (SSMs) map an input sequence $x(t)$ to an output $y(t)$ through a hidden state $h(t)$ \cite{gu2021combining, gu2021efficiently}. 
The Selective SSM (S6) mechanism underlying Mamba introduce input-dependent discretization:
\begin{align}
  \bar{A} &= e^{\Delta A}, & \bar{B} &= \Delta B, \\
  h_t &= \bar{A} h_{t-1} + \bar{B} x_t, & y_t &= C h_t + D x_t,
\end{align}
where $\Delta$, $B$, and $C$ are conditioned on the input.
To extend SSMs to 2D visual data, the SS2D module unfolds feature maps along multiple spatial routes (cross-directional scan) and fuses the outputs to obtain global receptive fields with linear complexity.

The Visual State Space (VSS) block builds upon SS2D within a residual framework, consisting of an SS2D-based spatial modeling branch and a feed-forward network branch for channel-wise transformation. 
Stacked VSS blocks form the VMamba backbone.
\section{Methods}
\subsection{Overview} % 4.1
We propose SCDM, an asymmetric dual-branch architecture designed to separate pathological features from normal anatomy under image-level supervision. Unlike standard fusion models, SCDM enforces uniquely specialized roles, as illustrated in Fig. 1: the Negative Branch actively suppresses normal anatomical structures (e.g., ribs), enabling the Positive Branch to focus entirely on localizing lesions. This disentanglement is achieved through an asymmetric strategy combining cooperative branch addition during training with competitive differential readout during inference.

% Overview figure
\begin{figure}[H]
  \centering
  \includegraphics[width=\linewidth, trim=0 7cm 0 7cm, clip]{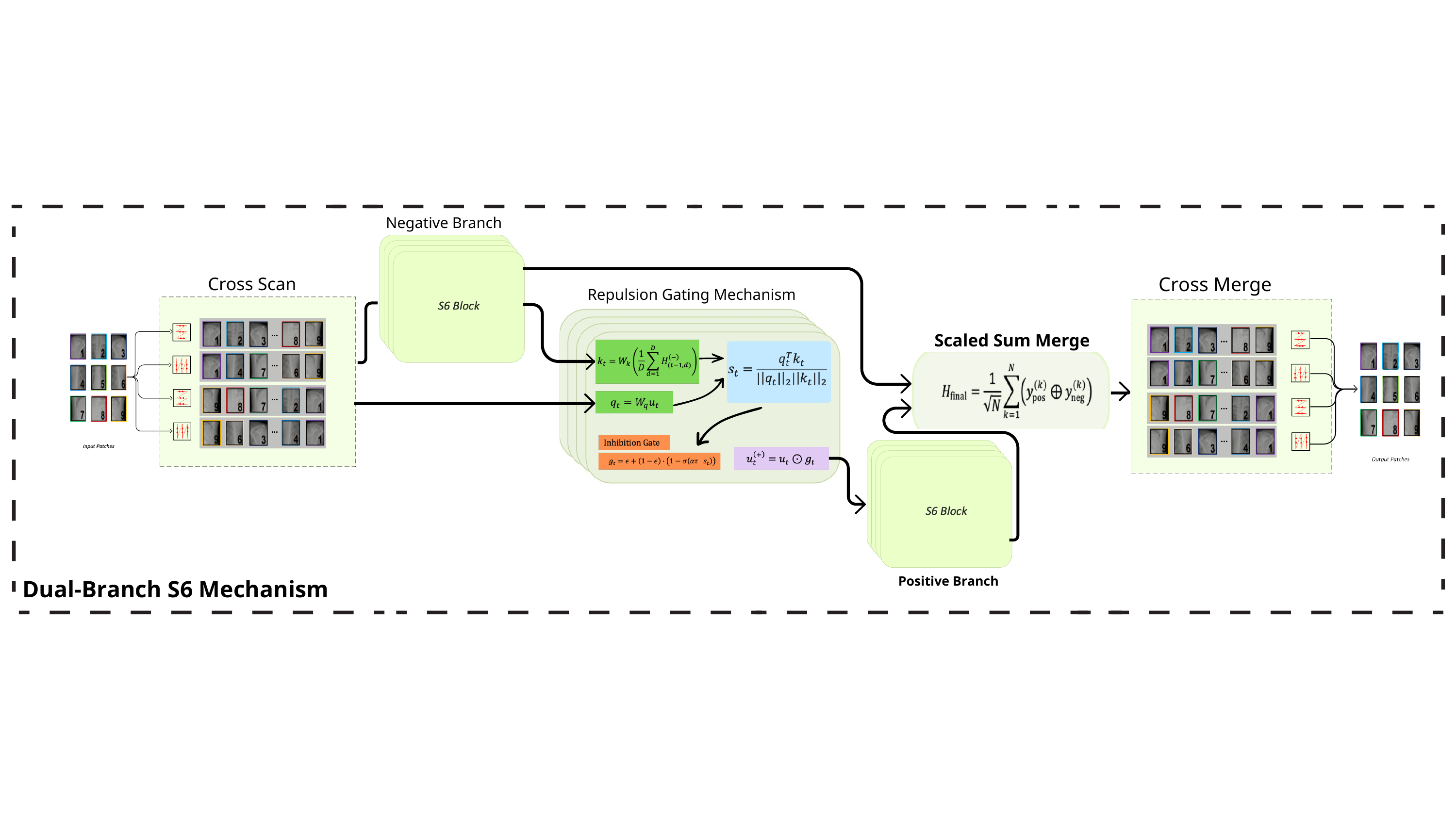}
  \caption{Overview of the proposed Dual-Branch S6 mechanism and the comparative memory interaction scheme within the VSS block.}
  \label{fig:scdm_overview}
\end{figure}

\subsection{Dual-Branch State Space Formulation}
The proposed SCDM architecture extends standard SSMs    by introducing two parallel branches with complementary objectives: a Negative Branch, denoted as $h_{\mathrm{neg}}$, for modeling normal anatomical context, and a Positive Branch, denoted as $h_{\mathrm{pos}}$, for detecting pathological deviations.

Both branches follow the continuous-time state space formulation:
\begin{equation}
  h'(t) = Ah(t) + Bx(t), \qquad y(t) = Ch(t).
  \label{eq:ct_ssm}
\end{equation}
Following discretization \cite{gu2021efficiently}, the recurrent updates for the dual-branch system are defined as:
\begin{align}
  h^{\mathrm{neg}}_t &= \bar{A}_{\mathrm{neg}}\, h^{\mathrm{neg}}_{t-1} + \bar{B}_{\mathrm{neg}}\, u_t, \\
  h^{\mathrm{pos}}_t &= \bar{A}_{\mathrm{pos}}\, h^{\mathrm{pos}}_{t-1} + \bar{B}_{\mathrm{pos}}\, (u_t \odot g_t),
  \label{eq:dual_branch_rec}
\end{align}
where $u_t \in \mathbb{R}^D$ represents the input token at step $t$, and $\bar{A}_{\mathrm{neg}}, \bar{B}_{\mathrm{neg}}, \bar{A}_{\mathrm{pos}}, \bar{B}_{\mathrm{pos}}$ are learnable transition and input matrices specific to each branch.

A key distinction of SCDM is that the Positive Branch receives a dynamically modulated input through a Repulsion Gate $g_t$, produced by the proposed repulsion mechanism. This design enables selective suppression of features already explained by the Negative Branch.

The final output of the SSM block is given by:
\begin{equation}
  y_t = (C_{\mathrm{neg}} h^{\mathrm{neg}}_t + D_{\mathrm{neg}} u_t) + (C_{\mathrm{pos}} h^{\mathrm{pos}}_t + D_{\mathrm{pos}} u_t).
  \label{eq:ssm_output_sum}
\end{equation}
This formulation enables disentangled yet complementary representation learning between anatomical context and pathological features.

\subsection{Repulsion Gating Mechanism}
To promote complementary feature learning, we introduce a similarity-driven \emph{Repulsion Gate} that dynamically suppresses the Positive Branch based on the historical hidden state of the Negative Branch. Unlike conventional gating mechanisms that rely solely on the current input, the proposed gate explicitly incorporates negative branch memory.
First, the current input is projected into a query vector:
\begin{equation}
  q_t = W_q u_t,
  \label{eq:repulsion_query}
\end{equation}
and the previous hidden state of the Negative Branch is aggregated across the feature dimension and projected into a key vector:
\begin{equation}
  k_t = W_k \left( \frac{1}{D} \sum_{d=1}^{D} H^{(-)}_{(t-1,d)} \right),
  \label{eq:repulsion_key}
\end{equation}
where $W_q \in \mathbb{R}^{D \times D}$ and $W_k \in \mathbb{R}^{D \times N}$ are learnable projection matrices, with $N$ denoting the state dimension and $D$ the feature dimension. Here, the aggregation operation computes a global state context by averaging the hidden states across all feature channels.
The similarity between the current input and the negative branch memory is computed using cosine similarity $s_t = (q_t^\top k_t) / (\lVert q_t \rVert_2 \lVert k_t \rVert_2)$. This similarity score is transformed into a gating coefficient via a temperature-scaled sigmoid:
\begin{equation}
  g_t = \epsilon + (1-\epsilon)\Big(1-\sigma(\alpha \tau s_t)\Big),
  \label{eq:repulsion_inhibition}
\end{equation}
where $\sigma(\cdot)$ denotes the sigmoid function, $\tau$ is a learnable temperature parameter controlling repulsion strength, $\alpha$ is a scaling constant (set to $10$ in practice), and $\epsilon$ is a stability constant (e.g., $0.05$) preventing complete signal suppression. Finally, the Positive Branch input is modulated as:
\begin{equation}
  u^{\mathrm{pos}}_t = u_t \odot g_t.
  \label{eq:repulsion_u_pos}
\end{equation}
This mechanism selectively suppresses hard negative mimics (e.g., bony structures) encoded by the Negative Branch, enabling the Positive Branch to isolate pathology-relevant features. To ensure optimization stability, we employ a curriculum learning strategy: the repulsion gate is disabled ($g_t = 1$) during an initial warm-up phase. This delay allows hem branches to establish foundational representations before introducing similarity-driven competition. Consequently, the Negative Branch first learns robust contextual anchors, which subsequently guide the Positive Branch toward specialized, complementary feature extraction.

\subsection{Integration into Vision Selective Scan (VSSc) Blocks}
The proposed dual-branch formulation is integrated into VSSc blocks. In intermediate layers, the outputs of both branches are merged to maintain architectural compatibility. However, in the final VSSc block, the branches are kept separate to preserve disentangled representations.

The final outputs are processed through independent feed-forward networks:
\begin{align}
  z_{\mathrm{neg}} &= \mathrm{FFN}_{\mathrm{neg}}\big(\mathrm{Norm}(y_{\mathrm{neg}})\big), \\
  z_{\mathrm{pos}} &= \mathrm{FFN}_{\mathrm{pos}}\big(\mathrm{Norm}(y_{\mathrm{pos}})\big).
  \label{eq:vss_integration_ffn}
\end{align}
These disentangled representations are pooled and fed into the classification head, where prediction is based on their competitive difference:
\begin{equation}
  y_{\mathrm{pred}} = \sigma\Big(\mathrm{Linear}(z_{\mathrm{pos}}) - \mathrm{Linear}(z_{\mathrm{neg}})\Big).
  \label{eq:vss_integration_pred}
\end{equation}
This competitive formulation encourages representation specialization and reduces feature overlap between contextual and pathology-sensitive branches.

\paragraph{Classification Loss ($L_{\mathrm{cls}}$).} The primary objective is binary cross-entropy (BCE) with label smoothing. During training, the prediction logits from both branches are added ($\text{logit}_{\text{pos}} + \text{logit}_{\text{neg}}$) to encourage cooperative signal accumulation:
\begin{equation}
  L_{\mathrm{cls}} = \mathrm{BCE}\Big(\sigma(\mathrm{logit}_{\mathrm{pos}} + \mathrm{logit}_{\mathrm{neg}}), y\Big).
\end{equation}
However, during inference, we apply a differential rule ($\text{logit}_{\text{pos}} - \text{logit}_{\text{neg}}$) to enforce a competitive decision margin.

\paragraph{Competitive Separation Loss ($L_{\mathrm{comp}}$).} To enforce strict branch specialization, this objective computationally maximizes the decision margin boundary between the Positive and Negative branches:
\begin{equation}
  L_{\mathrm{comp}} = -\log\Big(\sigma\big(\beta\, \tilde{y}\,(\mathrm{logit}_{\mathrm{pos}} - \mathrm{logit}_{\mathrm{neg}})\big)\Big),
\end{equation}
where $\tilde{y} = 2y - 1$. This loss actively pushes their representations apart for both diseased and healthy samples.

\paragraph{Safety Mechanisms ($L_{\mathrm{ortho}}$ and $L_{\mathrm{act}}$).} To reduce feature redundancy, we promote representational orthogonality ($L_{\mathrm{ortho}}$) between the Positive and Negative memory vectors exclusively for diseased samples, where separation between pathology and context is most critical: $L_{\mathrm{ortho}} = \frac{1}{N_{\mathrm{pos}}} \sum_{y_i=1} \big| (z_{\mathrm{pos}} \cdot z_{\mathrm{neg}}) / (\lVert z_{\mathrm{pos}} \rVert \lVert z_{\mathrm{neg}} \rVert) \big|$. Additionally, because strong repulsion gating may suppress the Positive branch excessively, we introduce an activity preservation loss ($L_{\mathrm{act}}$) that enforces a minimum activation norm: $L_{\mathrm{act}} = \frac{1}{N_{\mathrm{pos}}} \sum_{y_i=1} \mathrm{Softplus}( \mathrm{margin} - \lVert z_{\mathrm{pos}} \rVert )$. This prevents trivial zero-activation collapse and ensures stable gradient propagation. 

Collectively, model training is guided by the hybrid objective:
\begin{equation}
  L_{\mathrm{total}} = \lambda_{\mathrm{cls}} L_{\mathrm{cls}} + \lambda_{\mathrm{comp}} L_{\mathrm{comp}} + \lambda_{\mathrm{ortho}} L_{\mathrm{ortho}} + \lambda_{\mathrm{act}} L_{\mathrm{act}}.
\end{equation}

\section{Experiments}
We evaluated SCDM in the RSNA pneumonia data set \cite{stein2018rsna}, which consists of 20,672 normal and 6,011 pneumonia images. We use an 80/20 patient-wise split and resize all images to 128×128 using bilinear interpolation. The SCDM backbone uses a stage depth configuration of (1,1,1,1), whereas the VMamba-B baseline uses the standard (2,2,9,2) configuration. Our framework includes CLAHE preprocessing \cite{reza2004clahe} and various augmentations \cite{chlap2021review} (flipped/affine transforms, brightness/contrast adjustments, Gaussian noise, and coarse dropout) with a WeightedRandomSampler to address class imbalance.

\subsection{Comparison with Accuracy Metrics}

As shown in Table 1, we evaluated SCDM against representative convolutional, transformer-based, and state space architectures under identical training conditions to assess both discriminative performance and computational efficiency. The results indicate that the proposed dual-memory mechanism achieves competitive accuracy while providing a more efficient representation.

\begin{table}[H]
\caption{Quantitative comparison of performance and efficiency across CNN-based, transformer-based, and state space models.}
\label{tab:performance_summary}
\centering
\footnotesize
\renewcommand{\arraystretch}{1.1}
\setlength{\tabcolsep}{4pt}
\begin{tabular}{l c c c c c}
\hline
Model & Recall (\%) & Specificity (\%) & AUC & Params (M) & FLOPs (G) \\
\hline
Vmamba-B & 82.9 & 70.0 & 0.838 & 87.9 & 5.03 \\
Swin-B & 82.7 & 68.8 & 0.835 & 87.0 & 5.03 \\
ViT-Base Patch16 & 76.39 & 65.24 & 0.768 & 85.8 & 5.75 \\
ResNet-101 & \textbf{86.2} & 72.50 & \textbf{0.869} & 42.5 & 2.55 \\
SCDM (Proposed) & 78.7 & \textbf{77.0} & 0.858 & \textbf{29.4} & \textbf{1.44} \\
\hline
\end{tabular}
\end{table}

SCDM achieves competitive AUC with substantially lower recall than ResNet-101, and outperforms Swin-B, ViT-Base, and Vmamba-B, requiring only 29.4M parameters and 1.44 GFLOPs.

\begin{figure}[H]
\centering
\begin{subfigure}{0.48\textwidth}
    \centering
    \includegraphics[width=\linewidth]{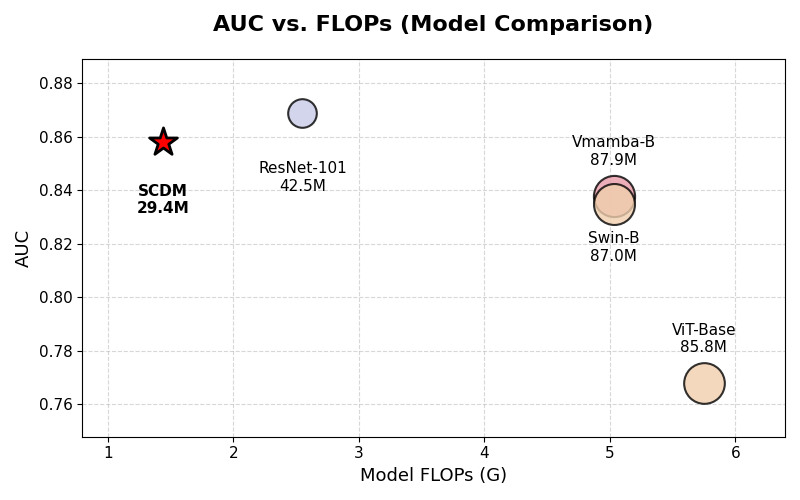}
    \caption{FLOPs comparison}
    \label{fig:flops}
\end{subfigure}
\hfill
\begin{subfigure}{0.48\textwidth}
    \centering
    \includegraphics[width=\linewidth]{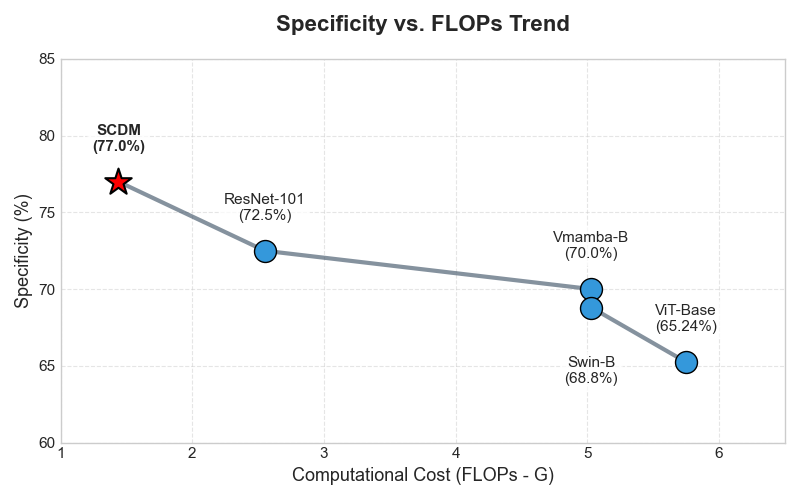}
    \caption{Model specifications}
    \label{fig:spec}
\end{subfigure}
\caption{Comparison of computational cost and model specifications.}
\label{fig:two_png_side_by_side}
\end{figure}

As shown in Table 1 and Fig. 2, SCDM maintains strong specificity while requiring substantially fewer parameters and FLOPs, indicating that the dual-memory mechanism enhances true negative discrimination without increasing model capacity. While overall metrics remain competitive, SCDM exhibits superior spatial selectivity, as analyzed in the next section.

\subsection{Representation-Level Analysis and Localization Behavior}

To further investigate the spatial behavior of the proposed dual-memory mechanism, we visualize activation maps using Grad-CAM \cite{selvaraju2017gradcam}. Fig. 3 compares the localization patterns of SCDM and representative baseline architectures.

\begin{figure}[H]
  \centering
  \includegraphics[
    width=0.8\linewidth,      % Görseli büyütür
    trim=13cm 6cm 14cm 7cm,     % left bottom right top
    clip
  ]{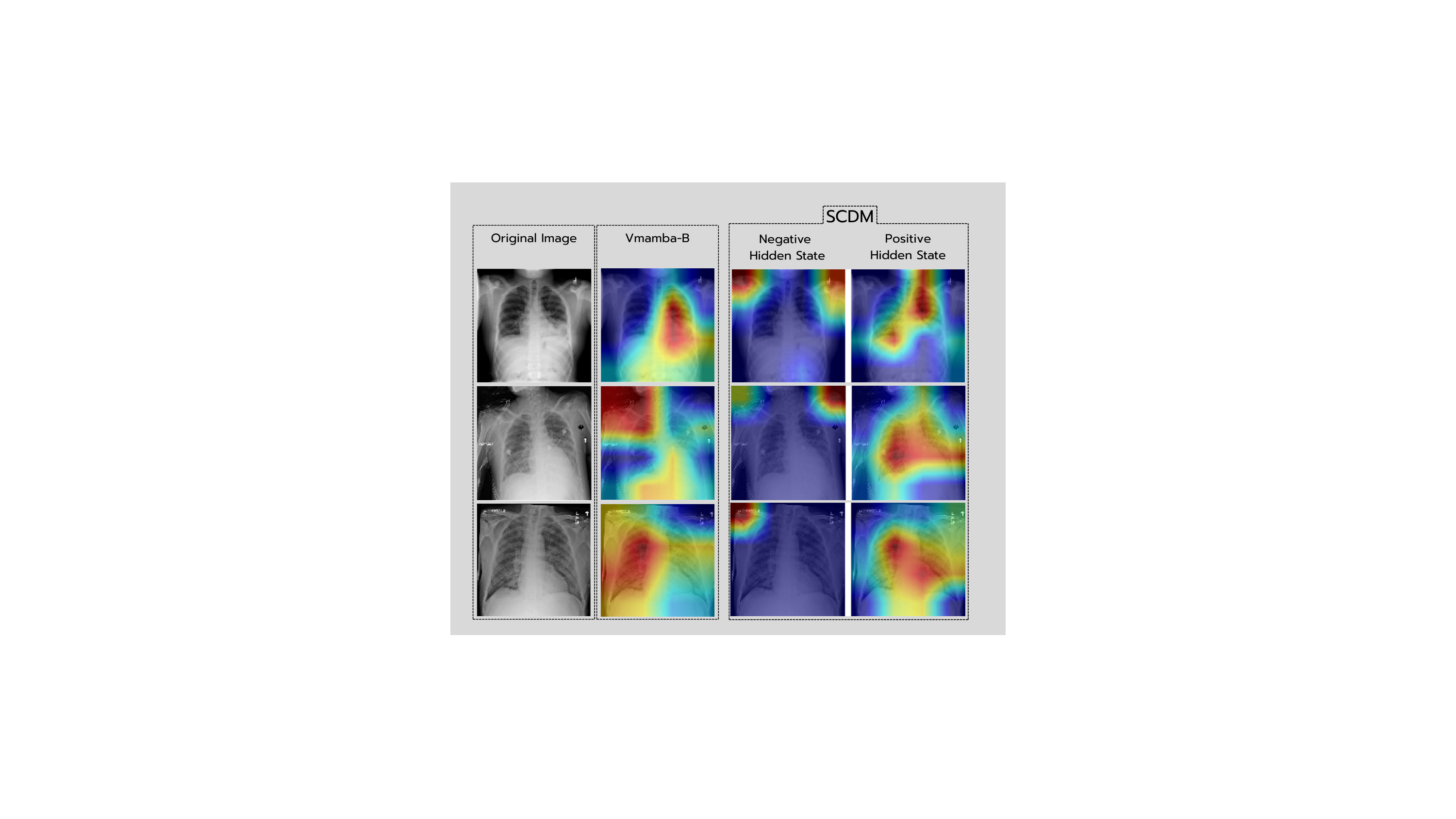}
  \caption{Grad-CAM visualization comparison between Vmamba and SCDM architectures.}
  \label{fig:scdm_gradcam}
\end{figure}
SCDM produces highly precise activation patterns compared to the diffuse maps of VMamba. This confirms the Negative Branch acts as an active anatomical suppressor, refining the Positive Branch's focus through differential inference. SCDM achieves this superior selectivity and interpretability with a lower computational footprint, validating the efficiency of our disentangled architecture.

\subsection{Component and Training Analysis}

To evaluate our asymmetric optimization, Table 2 compares different training objectives while keeping the differential readout ($P-N$, where $P$ and $N$ denote $\mathrm{logit}_{\mathrm{pos}}$ and $\mathrm{logit}_{\mathrm{neg}}$ respectively) fixed during inference. Purely competitive training (BCE($P-N$)) limits the network's ability to establish robust contextual anchors, degrading overall Recall. Conversely, our cooperative training objective (BCE($P+N$)) allows the Negative Branch to first act as a scaffold to capture anatomical distractors. When paired with the differential readout, this strategy significantly boosts Recall, enabling the Positive Branch to specialize purely on pathology. Figure 4 visually corroborates this: our asymmetric approach yields highly selective localization compared to standard fused representations.

\begin{table}[!htbp]
  \centering
  \caption{Comparing training objectives reveals that cooperative addition ($P+N$) provides essential scaffolding for the subsequent differential inference.}
  \label{tab:trainobj_inference_logic}
  \footnotesize
  \setlength{\tabcolsep}{5pt}
  \renewcommand{\arraystretch}{1.0}
    \begin{tabular}{l l c c c}
      \hline
      Train obj & Inference Logic & Recall (\%) & Specificity (\%) & AUC \\
      \hline
      BCE($P-N$) & $P-N$ & 77.8 & 77.1 & 0.852 \\
      \textbf{BCE($\mathbf{P+N}$) (Proposed)} & \textbf{$\mathbf{P-N}$} & \textbf{78.7} & \textbf{77.0} & \textbf{0.858} \\
      \hline
    \end{tabular}%
\end{table}

Separate analysis of the Dual-Branch architecture and Repulsion Mechanism reveals that while the dual structure provides capacity for feature separation, the Repulsion Gate drives sensitivity. By penalizing anatomical overlaps, it shifts the operating point towards high Recall (~5\% gain), prioritizing the minimization of false negatives in clinical screening despite a marginal trade-off in precision. Consequently, our fully integrated SCDM configuration offers the most robust balance for complex medical image classification.

\begin{figure}[H]
  \centering
  \includegraphics[
    width=0.85\linewidth,      % Görseli büyütür
    trim=14cm 10cm 14cm 11cm,     % left bottom right top
    clip
  ]{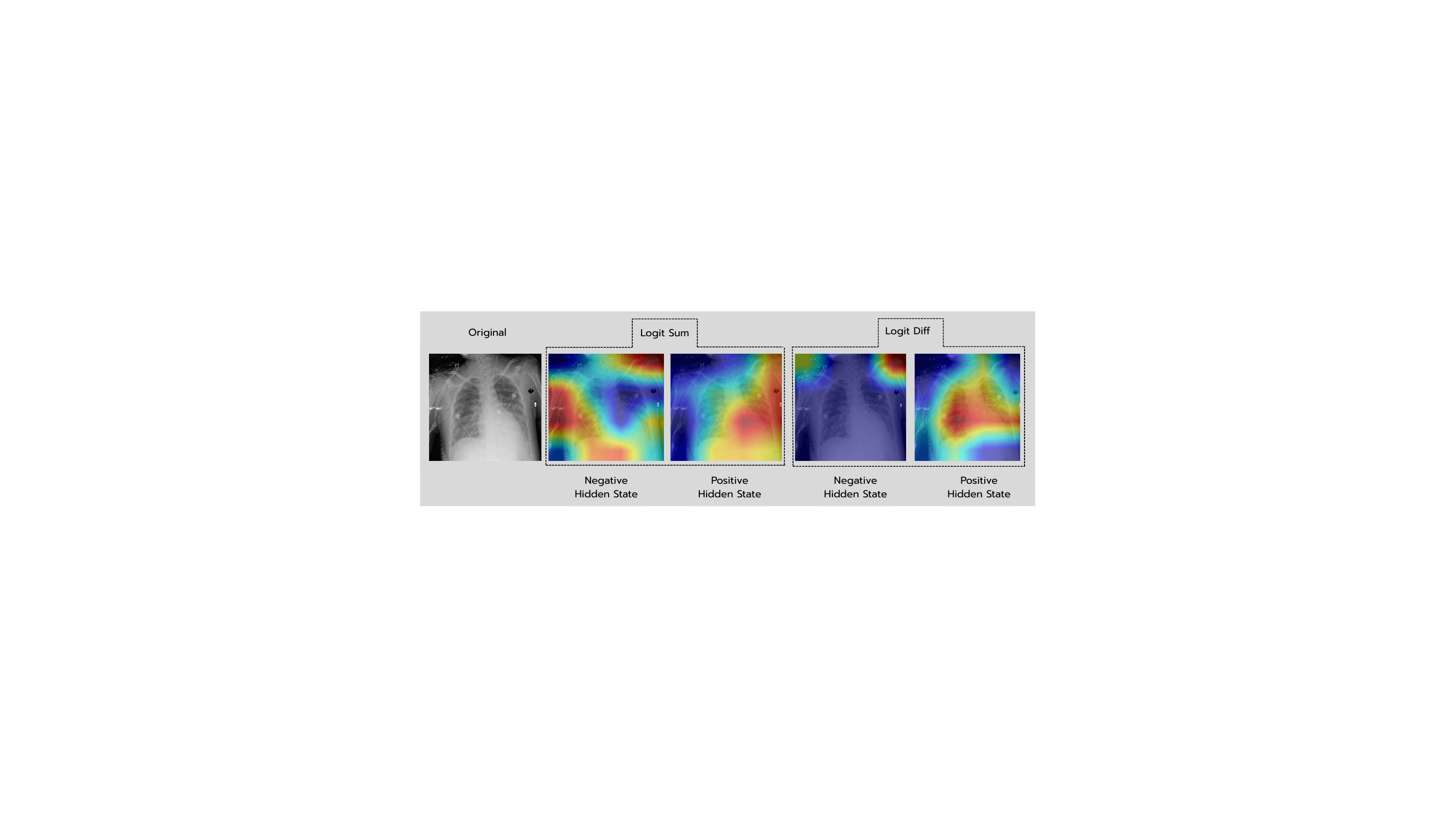}
  \caption{Grad-CAM visualization comparison between Training Dynamics.}
  \label{fig:logit_sum}
\end{figure}

\section{Conclusion}
We propose SCDM, leveraging dual-branch differential inference to disentangle focal pathology from anatomical distractors. While significantly improving sensitivity and localization by suppressing hard negatives,while requiring substantially fewer parameters and lower computational cost, demonstrating superior efficiency and scalability. Future work targets multi-class, high-resolution scaling.

\end{document}